\documentclass[journal]{IEEEtran}
\ifCLASSINFOpdf
\usepackage[pdftex]{graphicx}
\graphicspath{{../pdf/}{../jpeg/}}
\DeclareGraphicsExtensions{.pdf,.jpeg,.png}
\else
\usepackage[dvips]{graphicx}
\graphicspath{{../eps/}}
\DeclareGraphicsExtensions{.eps}
\fi

\usepackage{amsmath}
\def\BibTeX{{\rm B\kern-.05em{\sc i\kern-.025em b}\kern-.08emT\kern-.1667em\lower.7ex\hbox{E}\kern-.125emX}}

\usepackage{xspace}
\usepackage{url}
\usepackage{threeparttable}
\usepackage{dcolumn}
\usepackage{amsfonts}
\usepackage{color}
\newcolumntype{d}[1]{D{.}{.}{#1}}

\newcommand{\eat}[1]{}
\newcommand{\paratitle}[1]{\vspace{1ex}\noindent \textbf{#1}}

\let\oldhat\hat
\renewcommand{\vec}[1]{\mathbf{#1}}
\renewcommand{\hat}[1]{\oldhat{\mathbf{#1}}}
\renewcommand{\matrix}[1]{\mathbf{#1}}

\newcommand{\eg}{\emph{e.g.,}\xspace}
\newcommand{\rf}{\emph{rf.}\xspace}

\newcommand{\ie}{\emph{i.e.,}\xspace}
\newcommand{\etc}{\emph{etc.}\xspace}
\newcommand{\etal}{\emph{et al.}\xspace}
\newcommand{\aka}{\emph{a.k.a.,}\xspace}
\begin{document}
\title{Stack-VS:~Stacked Visual-Semantic Attention for Image Caption Generation}

\author{Wei~Wei,
    Ling~Cheng,
    Xianling~Mao,
    Guangyou Zhou,
    and Feida Zhu.
}

\markboth{IEEE Transactions on XXXX,~Vol.~X, No.~X, XX~XXXX}%
{Shell \MakeLowercase{\textit{et al.}}: Bare Demo of IEEEtran.cls for Journals}

\maketitle

\begin{abstract}
Recently, automatic image caption generation has been an important focus of the work on \emph{multimodal} translation task. Existing approaches can be roughly categorized into two classes, \ie \emph{top-down} and \emph{bottom-up}, the former transfers the image information (called as \emph{visual-level feature}) directly into a caption, and the later uses the extracted words (called as \emph{semantic-level attribute}) to generate a description.
However, previous methods either are typically based one-stage decoder or partially utilize part of \emph{visual}-level or \emph{semantic}-level information for image caption generation. In this paper, we address the problem and propose an innovative  multi-stage architecture (called as \textsf{Stack-VS}) for rich fine-gained image caption generation, via combining \emph{bottom-up} and \emph{top-down} attention models to effectively handle both \emph{visual}-level and \emph{semantic}-level information of an input image. Specifically, we also propose a novel well-designed stack decoder model, which is constituted by a sequence of decoder cells, each of which contains two LSTM-layers work interactively to re-optimize attention weights on both visual-level feature vectors and semantic-level attribute embeddings for generating a fine-gained image caption. Extensive experiments on the popular benchmark dataset \textsf{MSCOCO} show the significant improvements on different evaluation metrics, \ie the improvements on BLEU-4 / CIDEr / SPICE scores are $0.372$, $1.226$ and $0.216$, respectively, as compared to the state-of-the-arts.
\end{abstract}

\begin{IEEEkeywords}
	attention based mechanism, image captioning, novel object captioning.
\end{IEEEkeywords}

\IEEEpeerreviewmaketitle

\section{Introduction}
\label{sec:intro}


\IEEEPARstart{I}{mage} annotation has a significant effect for content-based image retrieval (CBIR)~\cite{socher2014grounded},
which is a process of assigning metadata in the form of captioning or keywords to an image.
However, manually image annotation is extremely expensive and time-consuming, especially since the dataset is really large and constantly growing in size.
Recently, \textbf{\underline{A}}utomatic \textbf{\underline{I}}mage \textbf{\underline{C}}aption \textbf{\underline{G}}eneration
(named \textsf{AICG}, \aka \emph{automatic image tagging})
has received a considerable amount of attention in computer vision (CV) and natural language processing (NLP) domain.
In fact, the \textsf{AICG} problem can be treated as a \emph{multi}-class image classification problem,
the goal of which is to automatically assign the annotations to the given image
via modeling of the correlations between the annotation words and the ``visual vocabulary'',~\ie extract visual features.

The challenge of \textsf{AICG} task lies in effectively modeling on both
\emph{visual}-level and \emph{semantic}-level information of the given image
for generating a meaningful human-like rich image description.
Inspired by machine translation,
great attention has been paid to exploit the encoder-decoder architecture
for image caption generation~\cite{25_,7_,24_,28_,4_,31_},
which commonly consists of a Convolutional Neural Network (CNN) based image feature  encoder
and a Recurrent Neural Network (RNN) based sentence decoder.
There already exist several efforts dedicated to research
on this topic, which can be roughly categorized into two classes,
\ie \emph{top-down}~\cite{24_,25_,7_,32_,1_} and \emph{bottom-up}~\cite{29_,23_}.
The former converts image information
(called as \emph{visual feature})
directly into descriptions,
while the later converts the generated words
(called as \emph{semantic attribute})
towards aspects of the given image into a word sequence.
%
%
However, these methods only partially address the challenges in \emph{visual}-level
or \emph{semantic}-level,
which still suffers from the following issues.
(i) Lack of end-to-end formulation that generates sentences based on individual aspects;
and (ii)  Generation of ambiguous information due to the whole image maps involved.
These drawbacks significantly reduce the accuracy and richness of the generated descriptions,
and naturally lead to the attempt that combines
such two models for caption generation
on both \emph{visual}-level and \emph{semantic}-level information.

Recently, there already have several works that employ a combination of
\emph{top-down} and \emph{bottom-up} models
to imitate human cognitive behavior through applying attentions to the salient image regions~\cite{2_,3_}.
However, most of such approaches are based on single stage training
and easily omit the interaction among detected objects during decoding, such as \cite{2_},
and thus predict error semantic attributes (\eg positional relations) among objects,
since ignoring the context-based fine-gained \emph{visual} information,
which inevitably leads to the difficulty in generating a rich fine-gained
description.

\begin{figure*}[!t]
	\centering
    \includegraphics[width=1.7\columnwidth, angle=0]{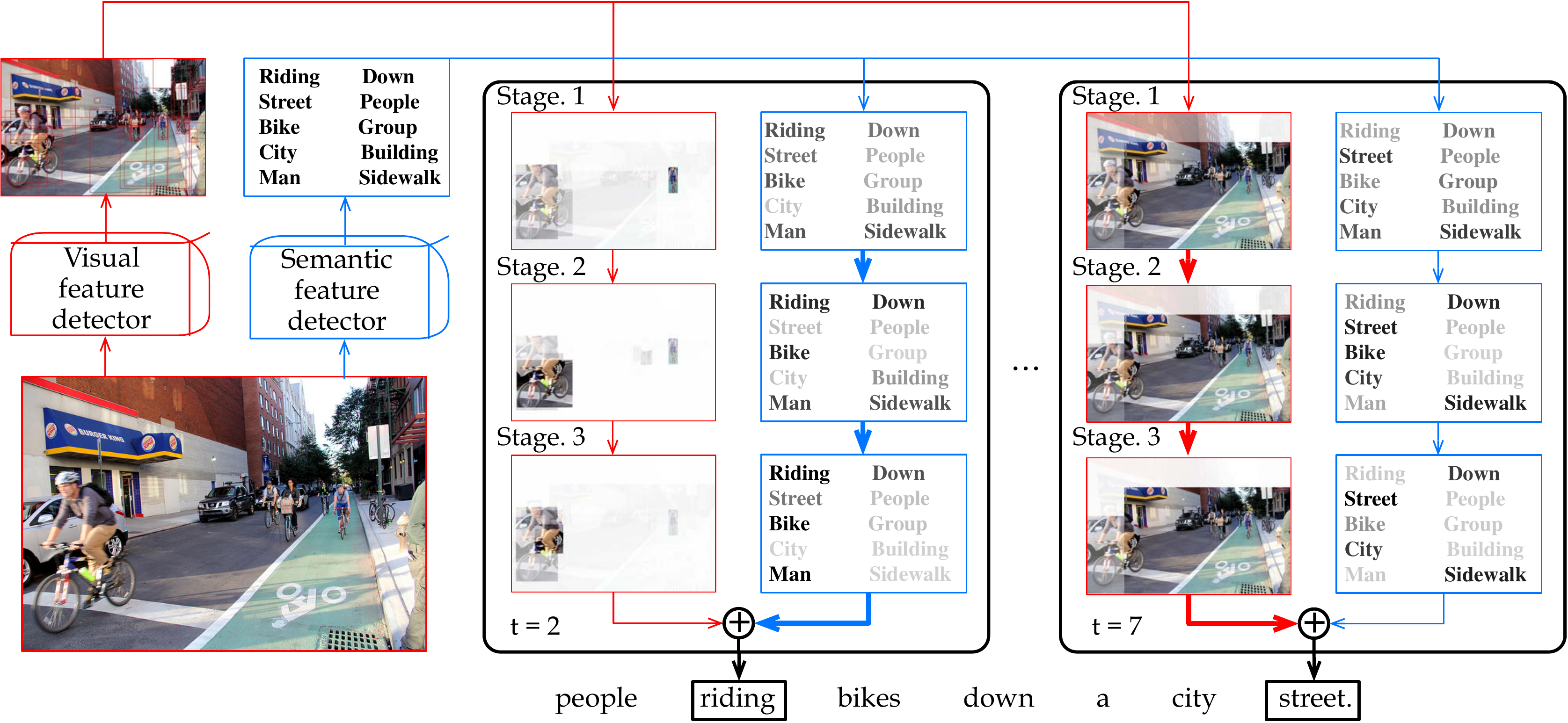}
    \caption{Illustration of our proposed coarse-to-fine framework. For each predicted word of the generated caption, the input image is firstly passed through a visual image detector and a semantic attribute detector, and then a series of image feature vectors and semantic attribute embeddings of salient regions are obtained.
    Subsequently, the proposed \textsf{Stack-VS} attention model gradually generates and refines the attention weights for such feature vectors and attribute embeddings at each time step, in which the thickness of  the red line and the blue line represent the contributions of the visual-level and semantic-level information derived from the input image to the predicted word of the generated caption through multi-stages. Consequently, the final word is the output of the final stage.
   }
	\label{fig:illustration_pipeline}
\end{figure*}

To this end, this paper proposes a unified \emph{coarse-to-fine} multi-stage architecture
to combine \emph{bottom-up} and \emph{top-down} approaches based on \emph{visual-semantic} attention model,
which is capable of effectively leveraging both the \emph{visual}-level image feature
and \emph{semantic}-level attributes for image caption generation.
Fig.~\ref{fig:illustration_pipeline} present an illustration regarding to our proposed method~(called as \textsf{Stack-VS}).
Specifically, there are two branches,~\ie a visual image detector and a semantic attribute detector,
to simultaneously construct the original visual-semantic information for salient regions during encoding,
and the decoder includes a stack of multiple decoder cells that are linked
for repeatedly generating the finer details,
of which each decoder cell consists of two \textsf{LSTM}~\cite{hochreiter1997long,donahue2015long}
attention-based layers,
namely,
\emph{visual-semantic} attention layer and a language model.
In particular, we employ the \emph{visual-semantic} attention layer to simultaneously
assign attention weights to visual features and semantic attributes,
and the attended image features and semantic attributes
are fed into a \emph{language} attention model for generating the hidden states of the current stage at each time step,
and then transferred to the next decoder cell for re-optimization.

The main contributions of this work are as follows.
%
\begin{itemize}
	
	\item We propose a novel \emph{coarse-to-fine} multi-stage architecture to combine \emph{bottom-up} and \emph{top-down} attention model
    for image caption generation, which is capable of effectively handling both visual-level and semantic-level information of the given image
    to generate a fine-gained image caption.
	\item We propose a well-designed stack model which links a sequence of decoder cells together, each of which contains two LSTM-layers work interactively
for generating fine-gained image caption via re-optimizing \emph{visual-semantic} attention weights during multi-stage decoding.
	\item Extensive experiments on popular benchmark dataset \textsf{MSCOCO} with two different optimizer, \ie \emph{ cross entropy}-based
and \emph{reinforcement learning}-based, which demonstrate the effectiveness of our proposed model, as compared to the state-of-the-arts, achieving BLEU-4 / CIDEr / SPICE scores are $0.372$, $1.226$ and $0.216$ respectively.	
\end{itemize}

\paratitle{Road Map}. The remaining of the paper is organized as follows. In Section \ref{sec:rel_work}, we review the related work. The proposed the learning of
coarse-to-fine image caption generation approach is given in Section \ref{sec:model}, followed by the experimental results in Section \ref{sec:exp}. Finally, Section~\ref{sec:con} concludes this paper.

\section{Related Work}
\label{sec:rel_work}

\paratitle{CNN+RNN based Model}.
With the success of \emph{sequence generation} in machine translation (\textsf{MT})~\cite{25_,24_},
\emph{encoder-decoder} model has been an important focus of the work on automatic image caption generation.
Mao \etal propose a two-layer factored neural network model~\cite{25_},
\ie a CNN-based \emph{visual}-level extraction network and a RNN-based \emph{word}-level embedding network,
for estimating the probability of generating the next word conditioned on the CNN extracted image feature and the previous word.
%
Analogous to~\cite{25_}, Vinyals \etal~ \cite{24_} propose a combination of CNN and RNN framework,
in which the visual features extracted by CNN are fed as the input at the first step,
to maximize a likelihood function of the target sentence.
Instead of giving the entire image features, \cite{7_} proposes an alignment model to learn a \emph{multimodal} embedding for generating the descriptions over image regions.
However, most of these models are inadequacy for the learning of RNN model owing to
making use of a fixed-size form to represent an image (\eg $4096$-dimensional CNN feature vector~\cite{6_}),
which might suffer from the imbalance problem when encoding-decoding the \emph{visual-semantic} information.

\paratitle{Attention-based model}.
There already exist several efforts dedicated to research on the \emph{automatic image caption generation} problem
by means of \emph{visual} attention mechanism~\cite{1_,18_,27_},
which steers the caption model to focus on the \emph{salient} image regions
when generating target words.
%
Recently, the existing attention-based approaches can be roughly categorized into two classes,
\ie \emph{bottom-up} and \emph{top-down}.
\emph{Bottom-up} approaches usually start with visual concepts, objects, attributes, words and phrases,
and then combine them into sentences by using \emph{language} model~\cite{2_}.
However, they may neglect the interaction among objects,
and thus these approaches may tend to maintain certain patterns to generate the final captions.
%
\emph{Top-down} approaches start with a ``gists'' of the image and convert them into several words~\cite{1_, 18_, 27_},
which enables the model to concentrate more on natural basis.
%
Although the great success has been achieved by \emph{top-down} approaches,
they cannot extract fine-gained image information from the given image due to mass noise data involved.

\paratitle{Fusion Model}.
To alleviate the above shortcomings, in fact many attempts have devoted to research on
how to improve the accuracy and richness of the generated descriptions through combining 
\emph{top-down} and \emph{bottom-up} architectures \cite{2_,3_,6_} together.
However, the vast majority of studies tend to transform the image information into object feature vectors or semantic word embeddings
to model the representations of the input image, which would lead to a noticeable decline in performance
due to the following facts.
First, unambiguous image information might also be introduced during encoding,
and directly compressing such information into image feature vectors would lead to the difficulty in the process of decoding semantic information
from the visual one;
Second, omitting the relations in visual-level and semantic-level of detected objects, \eg positional information,
would result in a unnatural caption generated.
Third, solely modeling the image caption problem through a one-stage training model is insufficient
for generating a fine-gained image.

Therefore, these methods still easily fail to achieve the image caption generation task.
Indeed, Gu \etal~\cite{gu2018stack} propose a multi-stage based framework for addressing 
the problem, in which the framework is constituted by multiple stages, 
and the predicted hidden states and the re-calculated attention weights
are repeatedly produced for generating the next word at each time step.
However, it cannot simultaneously handle \emph{visual-semantic} information during decoding,
and ignores the attention information from previous time steps.
In this work, our proposed approach is based on a \emph{coarse-to-fine} architecture with a stack of decoder cells
for repeatedly generating fine-gained caption via  multi-stages at each time step.
Specifically, each decoder cell consists of two LSTM layers that is capable of handling both \emph{visual}-level and \emph{semantic}-level information of the given image for better capturing the interactions of salient regions, as well as assigning different attention weights for visual-level and semantic-level information, respectively, which is benefit for generating more accurate and rich descriptions.

\section{Coarse-to-Fine Learning for Caption Generation}
\label{sec:model}
\begin{figure*}[!t]
	\centering
    \includegraphics[width=2.1\columnwidth, angle=0]{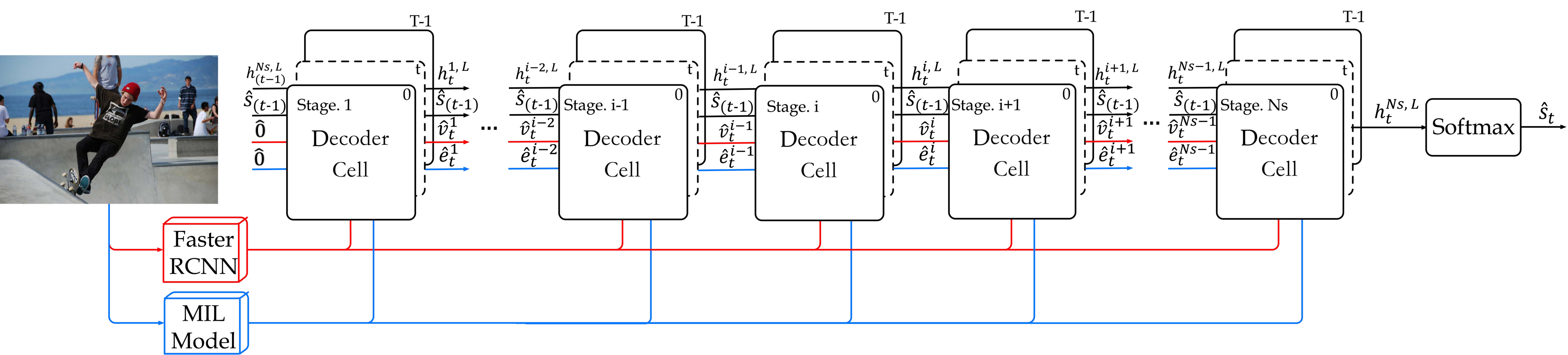}
	\caption{Overview of proposed approach, which is consists of a \emph{visual}-level image feature encoder~(Faster-RCNN), a \emph{semantic}-level attribute encoder~(MIL), as well as a stack of visual-semantic attention-based decoder cells.
Note that the red line denotes the visual-level branch and the blue one indicates the semantic-level branch, which are refined for generating the final captions via decoder cells within a ``stage-by-stage" structure.}
	\label{fig:overview_pipeline}
\end{figure*}

In this section,
we first present the formulation of our image caption generation problem in Section \ref{sec:method:pro_def},
followed by the illustration of image encoder in Section~\ref{sec:img_encoder}
and the coarse-to-fine decoder in Section~\ref{sec:coarse_2_fine}.
Then, the key structure of the decoder, \ie decoder cell,
is detailed
in Section~\ref{sec:method:decoder_cell},
and then we discuss the learning process regarding to our method.

\subsection{Problem Formulation \& Overview}
\label{sec:method:pro_def}
%
%
\paratitle{Problem Formulation}.~Given a pair-wise training instance $(I,\matrix{S})$,
%
$I$ and $\matrix{S} = \{s_i\}^{N_g}_{i=1}$ denote the input image
and its corresponding generated description sentence respectively,
where
$s_i$ is the $i$-th word of sentence $\matrix{S}$
and
$N_g$ is the sentence length.
Then, the caption generation problem can be formulated by employing a \emph{likelihood} function to minimize the \emph{negative} log
probability of the ground truth sentences for training,
which practically means we need to optimize
Eq. \eqref{eq:over_opt} over the entire training pairs via
an optimizer\footnote{Here, we use adaptive moment estimation (Adam~\cite{kingma2014adam}) for optimization.}
%
during training.
%
%
\begin{equation}
\label{eq:over_opt}
\begin{array}{ll}
\vec{\theta}^{*} \!\!&\leftarrow \mathop{\arg\min}_{\vec{\theta}}
{
\left(
-{
\underset{I\in \mathcal{I};\matrix{S}\in \mathcal{S}}{\sum}
\!\!\!\!\log \Pr(\matrix{S}|I;\vec{\theta})
}
\right)
}\\
&=\mathop{\arg\min}_{\vec{\theta}}
{
\left(
-{
\underset{I\in \mathcal{I};\matrix{S}\in \mathcal{S}}{\sum}
{\!\!\!\!\sum^{N_g}_{k=1}\log \Pr(s_{k}|s_{1:k-1},I;\vec{\theta})}
}
\right)
},
\end{array}
\end{equation}
where
$\Pr(s_{k}|I,s_{1:k-1})$ indicates the probability of generating the $k$-th word $s_{k}$
conditioned on the given image $I$ and the previous word sequence $s_{1:k-1}$;
$\vec{\theta}$ is the parameter set of the model and will be neglected for notational convenience;
$\mathcal{I}$ and $\mathcal{S}$ denote the training image set and its corresponding reference captions, respectively.
%
%
%
%

Therefore, the problem is transformed to estimate the
probability,
\ie
{\small
\begin{math}
  \Pr(s_{k}|s_{1:k-1},I).
\end{math}
}
Many approaches are proposed for this task~\cite{1_,3_,18_,2_,23_,31_}.
They are typically based on a deep RNN \emph{multi-modal} architecture in encoder-decoder manner,
the primary idea of which is to encode the visual feature vectors ($\matrix{V}$)~\cite{1_,3_,18_},
\emph{and/or} semantic attribute embeddings ($\matrix{E}$)~\cite{2_,23_,31_} into a
fixed-size context hidden state vector, and then decode it to generate a
possibly variably-sized description $\matrix{S}$.
Without loss of generality, Eq. \eqref{eq:over_opt} can be rewritten as,
\begin{equation}
\label{eq:over_opt_re}
- \sum_{I\in \mathcal{I};\matrix{S}\in \mathcal{S}}{\sum^{N_g}_{k=1}\log \Pr(s_{k}|s_{1:k-1},\matrix{V},\matrix{E})}.
\end{equation}


\paratitle{Overview}.~As aforementioned, the conventional approaches are commonly based on simple \emph{one-pass} attention mechanism~\cite{3_,5_,24_,mao2014deep},
and thus hardly generate rich and more human-like captions.
%
%
By following the previous work~\cite{gu2018stack},
we employ a \emph{coarse-to-fine } architecture to
exploit a sequence of \emph{intermediate}-level sentence decoders
for repeatedly refining image descriptions,
\ie
each stage predicts the incrementally refined description from the
preceding stage and passes it to the next one.
As such, let $\matrix{I^{(i-1)}}$ be the output of the preceding stage,
where $i \in [1,N_{s}]$ and $N_{s}$ denotes the total number of fine stages.
Consequently, Eq. \eqref{eq:over_opt_re} is converted by Eq. \eqref{eq:over_opt_final},
which means the sum of \emph{negative} log probability over the ground truth sentences,
\begin{equation}
\label{eq:over_opt_final}
-\sum_{I\in \mathcal{I};\matrix{S}\in \mathcal{S}}{\sum^{N_s}_{i=1}{\sum^{N_g}_{k=1}\log \Pr(s_{k}|s_{1:k-1},\matrix{V},\matrix{E},\matrix{I^{(i-1)}})}},
\end{equation}
where it should be noted that the first stage ($i\!=\!1$ ) is a coarse decoder that generates the coarse description~\cite{gu2018stack}, and the subsequent stages ($i\!>\!1$) are the attention-based fine decoders that increasingly produce the refined attentions for prediction.

\paratitle{Remark}.~By minimizing the loss function $\mathcal{L}$ in Eq. \eqref{eq:over_opt_final}, it enable to model the contextual relationships
among words in the target sentence.
The entire \emph{coarse-to-fine} framework
is depicted in Fig.~\ref{fig:overview_pipeline},
which mainly consists of two parts,
namely \textbf{\emph{image encoding}} and \textbf{\emph{coarse-to-fine decoding}},
and will be elaborated in detail later on.


\subsection{Image Encoding}
\label{sec:img_encoder}
%
Different from the previous work \cite{1_,4_,6_,18_,24_,30_,gu2018stack,GatedAttention}
that only encode the image in \emph{visual}-level,
in this paper we consider to simultaneously model the representation
of the given image $I$ from two different perspectives,
\ie \emph{visual}-level and \emph{semantic}-level, respectively.
%
%
%
More concretely, let $N_v$-dimensional vector
$\matrix{V}\!=\!\{\vec{v_{i}}\}^{N_v}_{i=1}$
be the \emph{visual}-level features of image $I$
and
$N_e$-dimensional vector
$\matrix{E}\!=\!\{\vec{e}_i\}^{N_e}_{i=1}~(\vec{e}_{i}\in \mathbb{R}^{d_e})$
be the \emph{semantic} attribute embedding
of image $I$ respectively,
which are obtained as follows,
\begin{itemize}
  \item \paratitle{Visual-level Image Feature}.
    To model the \emph{visual}-level features of the input image ($I$),
    we employ a widely-adopted approach, \ie
    \emph{Faster-RCNN} model~\cite{ren2015faster},
    to encode image ($I$) into a spatial feature vector~($\matrix{V}$)~\cite{1_,3_,18_}.
    %
    %
    Here,
    each spatial image feature $\vec{v_{i}}\in \mathbb{R}^{d_v}$  is a $d_v$-dimensional fixed-size feature vector (empirically set as $2048$)
    and denoted as the mean-pooled convolutional feature of the $i$-th salient region.
    It is encoded by CNN according to the detected bounding boxes with a \emph{bottom-up} attention mechanism and recursively activated at each time step.
    Additionally, each element $\vec{v_i}$ of the \textbf{\emph{initial}} visual feature vector (denoted as $\matrix{V_{0}}$) is generated based on the output of the final convolutional layer of \emph{Faster-RCNN}.

\item \paratitle{Semantic-level Attribute Embedding}.
An image ($I$) typically contains various semantic attributes
which is  paves a way to predict the probability distribution of semantic
instances over the massive attributes to obtain the
\textbf{\emph{initial}} list of \emph{semantic} attributes (denoted as $\matrix{E_{0}}$) that are most likely to appear in the image.
By following the work~\cite{31_}, we use a weakly-supervised multiple instance learning (called \textsf{MIL}~\cite{23_,2_}) to independently learn the most-likely \emph{semantic}-level attributes from the given image.
In particular, the part-of-speeches (POSs) of attributes in our proposed method are unconstrained,
\eg \emph{nouns}, \emph{verbs}, \emph{adjectives} and \etc,
which poses challenges in generating a much more natural captions,
and thus we further explore the correlations between the visual-level features and the semantic-level attributes of the given image~(\rf Section~\ref{sec:coarse_2_fine}).
For ease of implementation, each attribute is resized into
a $d_e$-dimensional fixed-size semantic attribute embedding ($\vec{e_i}$)
before decoding.
%
\end{itemize}



\begin{figure}[!t]
	\centering
    \includegraphics[width=1.0\columnwidth, angle=0]{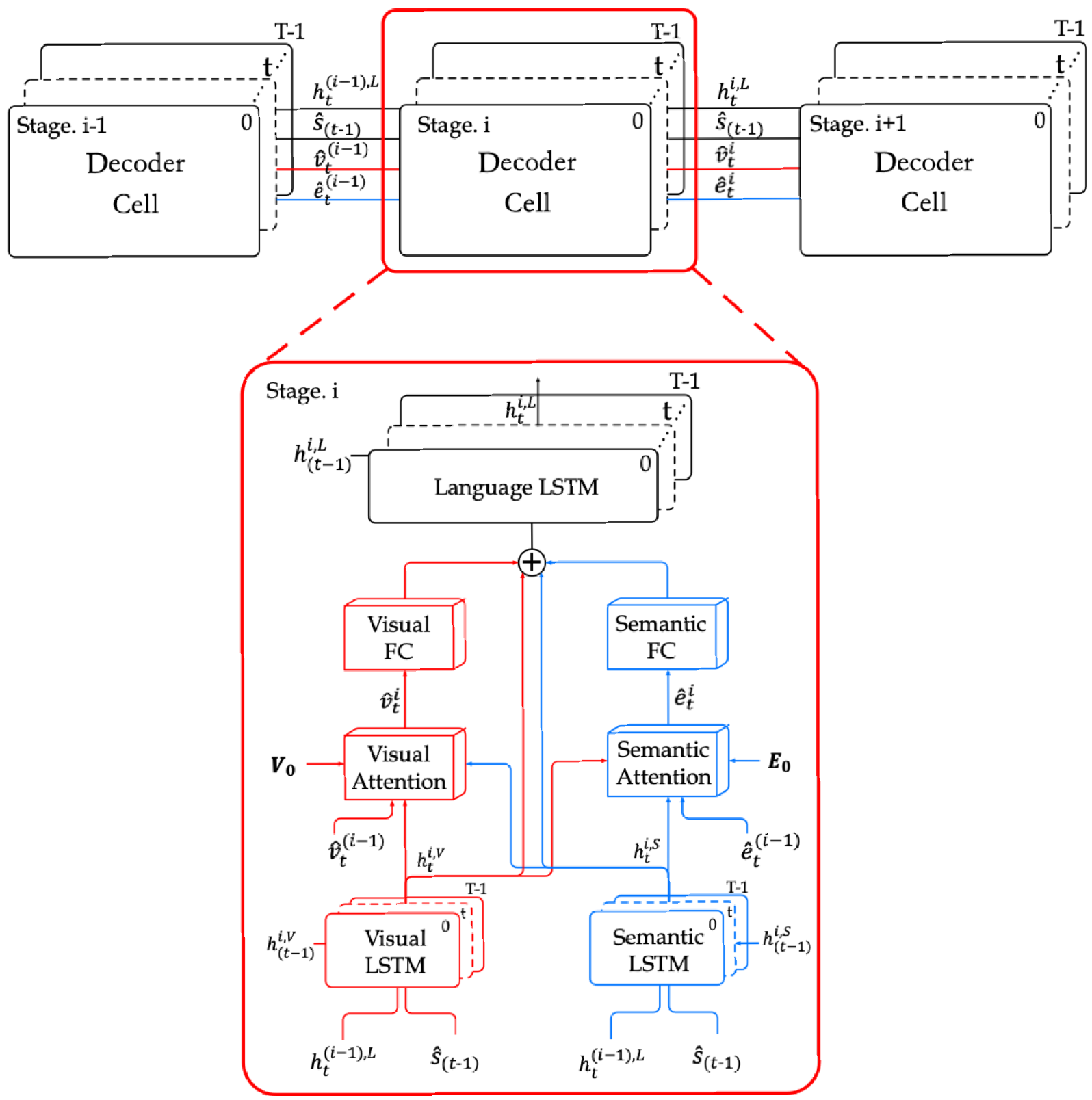}
    \caption{Structure of decoder cells, each of which is composed of two LSTM-layers to work interactively for generating attention weights on both \emph{visual}-level feature vectors and the \emph{semantic}-level attribute embeddings. Then, a language LSTM is employed to analyze the re-weighted visual-semantic information synthetically to successively generate the prediction word.}
	\label{fig:decoder_cell}
\end{figure}


\subsection{Coarse-to-Fine Decoding}
\label{sec:coarse_2_fine}
In this section,
we present our proposed \emph{coarse-to-fine} decoding architecture in detail,
%
which is composed of a sequence of attention-based decoders
to repeatedly produce the refined image descriptions.
%
For clarity, Fig.~\ref{fig:decoder_cell} presents the overall architecture of \emph{coarse-to-fine} sentence decoder.

More concretely, each predicted word of the generated description is decoded
within multiple stages at each time step $t\in [0,T-1]$, where $T$ is the length of the generated caption.
Without loss of generality,
let $N_{s}$ be the number of stages in our model,
where each stage is constituted by $T$ decoder cells,
%
%
in which each decoder cell is a predictor for producing the refined hidden state ($\vec{h^{i}_{t}}$) over a single time step $t$,
we will elaborate how to define each decoder cell later on~(\rf~\ref{sec:method:decoder_cell}).
Particularly, by following the work~\cite{gu2018stack},
we refer to the operation of the RNN unit (\ie \textsf{LSTM}~\cite{hochreiter1997long,donahue2015long})
to compute the hidden state of the $i$-th stage over a single time step $t$,
namely $\vec{h^{i}_{t}}$,
%
%
\begin{equation}
  \vec{h^{i}_{t}}=\mbox{\textbf{LSTM}}(\hat{s}_{(t-1)},\matrix{V_{0}}, \matrix{E_{0}},\matrix{I^{(i-1)}_{t}}),
\end{equation}
where
%
$\hat{s}_{(t-1)}$ denotes the predicted word embedding at time step $(t-1)$;
$\matrix{V}_{0}$ and $\matrix{E}_{0}$
indicate the \textbf{\emph{initial}} $N_v$-dimensional \emph{visual} feature vector
and $N_e$-dimensional \emph{semantic} attribute embeddings;
$\matrix{I^{(i-1)}_{t}}$ is the output of the the preceding stage,
which is defined as
\begin{equation}
\matrix{I^{(i-1)}_{t}} =  [\vec{h^{(i-1)}_{t}}, \hat{v}_t^{(i-1)}, \hat{e}_t^{(i-1)}]
\end{equation}%
where
$\hat{v}_t^{(i-1)}$ and $\hat{e}_t^{(i-1)}$ are the attended \emph{visual} feature vector
and the attended \emph{semantic} attribute embedding of the $(i-1)$-th stage at time step $t$.
The details of such stuffs will be elaborated in Section \ref{sec:method:decoder_cell}.

\paratitle{Remark}. In particular, the first stage of our model is called \textbf{\emph{coarse}} stage,
which generates a \emph{roughly} attention weighted caption
based on both the initial visual-level feature vector and the semantic-level attribute embedding.
%
Accordingly, we can obtain a \emph{coarse} prediction for the next stage,
where the corresponding input
is given by $(\hat{s}_{(t-1)}, \matrix{V_{0}}, \matrix{E_{0}}, \matrix{I^{0}_{t}})$,
and $\matrix{I^{0}_{t}}$ is defined as
\begin{math}
   [\vec{h}^{N_s}_{(t-1)}, \hat{v}^{0}_{t}, \hat{e}^{0}_{t}],
\end{math}
in which $\vec{h}^{N_s}_{(t-1)}$ is the hidden state of the final stage at time step $(t-1)$;
and $\hat{v}^{0}_{t}$ is a vector with all its elements equal to $0$~(similar for $\hat{e}^{0}_{t}$).

\subsection{Decoder Cell}
\label{sec:method:decoder_cell}
As mentioned in work~\cite{24_,gu2018stack}, modeling
of the entire information of an image for
each single target word generation would lead to
\emph{sub-optimal} solution due to the irrelevant
cues introduced from other salient regions.
Therefore, we need to calculate an attention based on much more natural basis
for generating  a more \emph{human}-like caption.

Inspired by \cite{24_} but \emph{\textbf{entirely different}} from their work, each decoder cell at time step $t$ is designed
as shown in Fig.~\ref{fig:decoder_cell},
in which the output of each decoder cell is generated through two \textsf{LSTM} layers,
\ie
\emph{visual-semantic} \textsf{LSTM} layer and \emph{Language} \textsf{LSTM} layer. Next, we will elaborate them in detail.

\paratitle{Visual-Semantic \textsf{LSTM} Layer}.
As shown in Fig.~\ref{fig:decoder_cell},
there are two different branches in \emph{visual-semantic} \textsf{LSTM} layer,
\ie the red line denotes the flow diagram of
\emph{visual} attention model~(denoted as $\mbox{LSTM}^{V}$),
and the blue one indicates the flow diagram of \emph{semantic} attention model~(denoted as $\mbox{LSTM}^{S}$),
which work interactively to calculate the attention weights on
\emph{visual} feature vectors and \emph{semantic} attribute embeddings.

\paratitle{Visual-Attention Model}.
To be specific,
the input of $\mbox{LSTM}^{V}$ is used to model the maximum \emph{visual}-level context
based on the output of Language \textsf{LSTM} of the \textbf{\emph{preceding}} stage,
as well as the predicted word embedding generated so far.
Hence, the input $\vec{x^{i,\matrix{V}}_{t}}$ to \emph{visual} attention model in the $i$-th stage at time step $t$
consists of the previous hidden state $\vec{h^{(i-1),L}_{t}}$ of the \emph{language} \textsf{LSTM},
concatenated with the predicted word embedding
at time step $(t-1)$, which is
\begin{equation}
  \vec{x^{i,\matrix{V}}_{t}} = [\hat{s}_{(t-1)},\vec{h^{(i-1),L}_{t}}],
\end{equation}

Then, the output of $\mbox{LSTM}^{V}$ can be computed as follows:
\begin{equation}
\vec{h^{i,V}_t} = \mbox{\textbf{LSTM}}^{V} (\vec{x^{i,V}_{t}}, \vec{h^{i,V}_{(t-1)}})
\end{equation}

To tackle the \emph{sub-optimal} problem,
we generate a visual attention weight for the $k$-th \emph{visual} feature vector $\vec{v_k}\in \matrix{V_0}$
of the $i$-th stage at time step $t$, which is given by
\begin{equation}
\begin{split}
{a}^{i,V}_{t,k}= \vec{W^V_{a}}\tanh(
\matrix{W}^V_{v,a}\vec{v_k}
+\matrix{W}^V_{\hat{v},a}\vec{{\hat{v}}_t^{(i-1)}}\\
+\matrix{W}^{V}_{h_v,a}\vec{h^{i,V}_t}
+\matrix{W}^{V}_{h_s,a}\vec{h^{i,S}_t}
),
\end{split}
\end{equation}%

\begin{equation}
[{\alpha}^{i,V}_{t,1}, \ldots, {\alpha}^{i,V}_{t,N_v}] =
\mbox{\textbf{softmax}}[{a}^{i,V}_{t,1}, \ldots, {a}^{i,V}_{t,N_v}],
\end{equation}%
where $\vec{W^V_{a}} \in \mathbb{R}^{1\times{d_a}}$,
$\matrix{W}^V_{v,a} \in \mathbb{R}^{{d_a}\times{d_v}}$,
$\matrix{W}^V_{\hat{v},a} \in \mathbb{R}^{{d_a}\times{d_v}}$
$\matrix{W}^{V}_{h_v,a}\in \mathbb{R}^{d_a\times{d_h}}$,
$\matrix{W}^{V}_{h_s,a}\in \mathbb{R}^{d_a\times{d_h}}$
are trainable parameter matrices;
and
$d_a$ is the number of hidden units in \emph{visual} (or \emph{semantic}) attention model;
$d_h$ is the dimension of the output of $\mbox{LSTM}^{V}$
(or $\mbox{LSTM}^{S}$).
Then, the attended visual feature vector can be calculated as follows:
\begin{equation}
\hat{v}^{i}_t =  \sum_{k=1,\vec{v_k}\in \matrix{V_0}}^{N_v}{\alpha}^{i,V}_{t,k}\vec{v_k}.
\end{equation}%

\paratitle{Semantic Attention Model}.
Analogous to visual attention model, we can easily obtain the output of $\mbox{LSTM}^{S}$ can be calculated as follows
\begin{equation}
\vec{h^{i,S}_t} = \mbox{\textbf{LSTM}}^{S} (\vec{x^{i,S}_{t}}, \vec{h^{i,S}_{(t-1)}}).
\end{equation}
%

For the $k$-th \emph{semantic} feature vector $\vec{e_k}\in \matrix{E_0}$
of the $i$-th stage at time step $t$, the semantic attention weight
is given by
\begin{equation}
\begin{split}
{a}^{i,S}_{t,k}= \vec{W^S_{a}}\tanh(
\matrix{W}^S_{e,a}\vec{e_k}
+\matrix{W}^S_{\hat{e},a}\vec{{\hat{e}}_t^{(i-1)}}\\
+\matrix{W}^{S}_{h_v,a}\vec{h^{i,V}_t}
+\matrix{W}^{S}_{h_s,a}\vec{h^{i,S}_t}
),
\end{split}
\end{equation}%
\begin{equation}
[{\alpha}^{i,S}_{t,1}, \ldots, {\alpha}^{i,S}_{t,N_e}] =
\mbox{\textbf{softmax}}[{a}^{i,S}_{t,1}, \ldots, {a}^{i,S}_{t,N_e}].
\end{equation}%

Consequently, the attended semantic attribute embedding can be calculated as follows:
\begin{equation}
\hat{e}^{i}_t =  \sum_{k=1,\vec{e_k}\in \matrix{E_0}}^{N_e}{\alpha}^{i,S}_{t,k}\vec{e_k}.
\end{equation}%
%

%

\paratitle{Language LSTM}.
Different to the work~\cite{3_},
we model the \emph{Language} model based on both the \emph{visual}-level
and \emph{semantic}-level information of the given image $I$.
Specifically,
%
the input of $\mbox{LSTM}^{L}$ is constituted by
the \emph{visual-semantic} information of the given image,
that is,
the sum of the outputs $(\vec{h^{i,V}_t},\vec{h^{i,S}_t})$ of $\mbox{LSTM}^{V}$ and $\mbox{LSTM}^{S}$,
as well as the outputs $(\hat{v}^{i}_t, \hat{e}^{i}_t)$ of the visual attention model
and the semantic attention model.
However, there exists the \emph{cross-modal} problem,
\ie the dimension of $\hat{v}^{i}_t$ is different from $\hat{e}^{i}_t$
since we utilize different approaches to initializing
the \emph{visual} feature vector ($\matrix{V_0}$)
and the \emph{semantic} attribute embeddings ($\matrix{E_0}$).
As such,
we use two fully connected layers, \ie $\matrix{FC^V}$, $\matrix{FC^S}$
to convert them into a unified form (the dimension of which is the same as $\vec{h^{i,V}_t}$ and $\vec{h^{i,S}_t}$), and thus
the input~($\vec{x^{i,L}_t}$) and the output ($\vec{h^{i,L}_t}$) of $\mbox{LSTM}^{L}$ is given by,
\begin{equation}
\vec{x^{i,L}_t} =  (\mbox{FC}^V(\hat{v}_t) + \mbox{FC}^S(\hat{e}_t) + \vec{h^{i,V}_t} + \vec{h^{i,S}_t}),
\end{equation}%
\begin{equation}
\label{eq:final_h_ilt}
\vec{h^{i,L}_t} =  \mbox{\textbf{LSTM}}^L (\vec{x^{i,L}_{t}}, \vec{h_{t-1}^{i,L}}),
\end{equation}%
where $\vec{h_{t-1}^{i,L}}$ denotes the the hidden state of last \emph{Language LSTM}.

Subsequently, the probability of generating the $k$-th word $s_k$
given
the previous word sequence $s_{1:k−1}$,
the initial visual vector features $\matrix{V_{0}}$,
and semantic attribute embedding $\matrix{E_{0}}$, are concatenated with the
output of the preceding stage $\matrix{I^{(i-1)}}$,
\ie $\Pr(s_{k}|s_{1:k-1},\matrix{V},\matrix{E},\matrix{I^{(i-1)}})$ in Eq. \eqref{eq:over_opt_final},
which can be rewritten as follows,
%
%
\begin{equation}
\label{eq:over_opt_final_res}
\Pr(s_{k}|s_{1:k-1},\matrix{V_0},\matrix{E_0},\matrix{I^{(i-1)}})
=\mbox{\textbf{softmax}}({\matrix{W^L_{h_l,p}}} \vec{h^{i,L}_k})
\end{equation}
where $\vec{h^{i,L}_k}$ denotes the hidden state of \emph{Language LSTM} given by Eq. \eqref{eq:final_h_ilt};
$\matrix{W}^{L}_{h_l,p}\in \mathbb{R}^{d_p\times{d_h}}$ refers to a trainable parameter matrix,
and $d_p$ is the vocabulary size.

\subsection{Learning}
Indeed, the \emph{coarse-to-fine} framework is a complex deep architecture,
its training process would easily leads to the \emph{vanishing gradient} problem
due to the noticeable decline in the magnitude of gradients when back-propagated
via multiple stages~\cite{gu2018stack}.
To tackle the problem, we consider to employ a cross-entropy (XE) loss to
incorporate ground-truth information into the learning process of intermediate layers for supervision,
and thus the overall loss function is defined
by cumulatively calculating of the cross-entropy (XE) loss at each stage,
which is computed given by Eq.~\eqref{eq:over_opt_final}
{
\begin{equation}
\label{eq:loss_xe}
  \hspace{-0.35cm}\mathcal{L}_{XE}(\theta)\!\!=\!\!
  -\!\!\!\!\!\!\sum_{I\in \mathcal{I};\matrix{S}\in \mathcal{S}}
{
\sum^{N_s}_{i=1}
\sum^{N_g}_{k=1}\log
\left(
p_{\theta_{0:i}}(s_{k}|s_{1:k-1},\matrix{V_0},\matrix{E_0},\matrix{I^{(i-1)}})
\right),
}
\end{equation}}
where
\begin{math}
  p_{\theta_{0:i}}(s_{k}|s_{1:k-1},\matrix{V_0},\matrix{E_0},\matrix{I^{(i-1)}})
\end{math}
is the probability of the $k$-th word $s_{k}$ given by the output of $\mbox{LSTM}^{L}$ decoder at the $i$-th stage;
and $\theta_{0:i}$ denotes the parameters up to the $i$-th stage decoder.
Note that the attended weights of our model are needed to be re-trained at each time step due to the received information to fluctuate at each stage basis ,
it is quite different from the shared weight network in~\cite{gu2018stack}.

However, the \emph{exposure bias} issue has not been fundamentally solved
in existing \emph{cross-entropy} loss based training approaches~\cite{gu2018stack,3_,5_}.
To this end, an alternative solution is introduced to treat the generative model
as a reinforcement learning based method~\cite{19_,20_,5_},
in which the \emph{Language LSTM} decoder
is viewed as an ``agent'' to interacts with an external ``environment''
(\ie the \emph{visual-semantic} information of the image).
For simplicity, let $\tilde{\matrix{S}}=\{\tilde{\vec{s}}_{t}\}_{1:N_T}$ be the
sampled caption and each word $s_{k}$ is sampled
from the output of the final stage ($N_s$) at time step $t$, according to an action $\vec{s_t}\!\!\sim\!\! p_{\theta}$,
where $p_{\theta}$ is a policy based on the parameterized network $\theta$,
and $N_T$ is the length of the sampled caption.
To minimize the negative expected rewards (punishments), the estimation is defined as
\begin{equation}
\label{eq:eva:r}
\begin{array}{ll}
\mathcal{L}_{RL}(\theta)
&\!\!\!\!=-\mathbb{E}_{\tilde{\matrix{S}}\sim p_{\theta}}
[r(\tilde{\matrix{S}})]\\
&\!\!\!\!\approx -r(\tilde{\matrix{S}}),~\tilde{\matrix{S}}\!\sim\!p_{\theta}
\end{array}
\end{equation}
where $r(.)$ is the reward function (\eg \textsf{CIDEr})
that is computed by comparing the generated caption with the corresponding
reference captions of the input image using the standard evaluation metric.

To reduce the variance of the gradient estimate, for each training sample,
the expected gradient can be approximated with a single Monte-Carlo sample
$\vec{s_t}\!\!\sim\!\! p_{\theta}$ by following the work~(Self-Critical Sequence Training, SCST~\cite{5_}),
which is computed by
\begin{equation}
\label{eq:loss_final}
{\nabla}{\mathcal{L}_{RL}(\theta)} \approx
-\left(r(\tilde{\matrix{S}})-\vec{b}\right)
{\nabla_\theta}\log{p_{\theta}}(\tilde{\matrix{S}}),
\end{equation}%
where $\vec{b}$ is a score function of $\theta$ that is empirically set as $\mbox{CIDEr}(\hat{\matrix{S}})$, and $\hat{\matrix{S}}$ is
the sampled reference caption by greedy decoding.

\paratitle{Remark}.
Essentially, the equation given in Eq.~\eqref{eq:loss_final} trends to increases the probability of sampled captions with higher scores
than the samples generated based on the current model.
\section{Experimental Results}
\label{sec:exp}
In this section, we first present the experimental datasets and the evaluation metrics,
and then describe the baselines for comparison with the data-processing and the implementation details.
Finally,  we conduct extensive experiments to evaluate the performance of our proposed algorithm.

\subsection{Datasets}
%
We evaluate our proposed method on a popular benchmark dataset
%
\ie \textsf{MSCOCO}~\cite{16_,17_}, which contains $164,062$ images with $995,684$ captions,
and each image has at least five reference captions.
The dataset is further divided into three parts, \ie \emph{training} ($82,783$ images),
\emph{validation} ($40,504$ images) and \emph{testing} ($40,504$ images),
which are withheld in \textsf{MSCOCO} server for \emph{on-line} comparison.
%
To evaluate the quality of the generated captions, we follow ``\emph{karpathy split}'' evaluation strategy in \cite{10_}, \ie $5,000$ images are chosen for \emph{offline} validation,
and another $5,000$ images are selected for \emph{offline} testing.
Then the evaluation results are reported by the publicly available \textsf{MSCOCO} Evaluation Toolkit\footnote{\url{https://github.com/tylin/coco-caption}.},
as compared to the state-of-the-arts listed in the leader board on \textsf{MSCOCO} online evaluation server\footnote{\url{http://cocodataset.org/#captions-eval}.}.
We also evaluate our proposed model \textsf{Stack-VS} on such platform.

\subsection{Evaluation Metrics}
%
To evaluate the image caption generation performance of different approaches,
we adopt the widely used evaluation metrics, \ie
\textbf{BLEU}, \textbf{Rouge}, \textbf{METEOR} and \textbf{CIDEr}.
In particular, we also adopt a widely used evaluation metric, \ie~{\bf SPICE},
which is more similar to human evaluation criteria.
For all the metrics, the higher the better.

\paratitle{BLEU}~\cite{11_}, which is widely used in machine translation domain, and defined as the geometric mean of n-gram precision scores multiplied by a brevity penalty for short sentences.

\paratitle{ROUGE}~\cite{12_}, which is initially proposed for summarization and is used to compare the overlapping n-grams, word sequences and word pairs.
Here, we use its variant, \ie \emph{ROUGE-L}, which basically measures the longest common subsequences between a pair of sentences.

\paratitle{METEOR}~\cite{9_}, which is a machine translation metric and is defined as the harmonic mean of precision and recall of uni-gram matches between sentences,
via making use of synonyms and paraphrase matching.

\paratitle{CIDEr}~\cite{10_}, which measures the consensus between candidate image description and the reference sentences
provided by human annotators.
Particularly, in order to calculate it, an initial stemming is applied and each sentence is represented with a set of $1$-$4$ grams.
Then, the co-occurrences of n-grams in the reference sentences and candidate sentence are calculated.

\paratitle{SPICE}~\cite{8_}, which is estimated according to
the agreement of the scene-graph tuples of the candidate sentence and all reference sentences.
Specifically, the scene-graph is essentially a semantic representation of parsing the input sentence to a set of semantic tokens,
\eg object classes, relation types or attributes.

\subsection{Baseline Methods \& Parameter Setting}

Here, we compare our approach with
To evaluate the effectiveness of our proposed method,
we compare our model with the state-of-the-art methods
as follows,

\begin{itemize}
	\item [1)]
	{\bf Hard-Attention}~\cite{1_}. This method incorporates spatial attention on convolutional features of an image into a encoder-decoder framework through
training ``hard'' stochastic attention by reinforce algorithm. Note that, this method also can be trained with ``soft'' attention mechanism with standard back-propagation, however we neglects it due to the poor performance reported as compared to the ``hard'' one.
	\item [2)]
	{\bf Semantic-Attention}~\cite{2_}. This method models selectively attend to semantic concept proposals, then mix them into the hidden states and the outputs of the recurrent neural networks for image caption generation. Essentially, the selection and fusion is regarded as a feedback that connects the computation of \emph{top-down} and \emph{bottom-up}.
	\item [3)]
	{\bf MAT}~\cite{6_}. This method converts the input image into a sequence of detected objects that feeds as the source sequence of the RNN model, and then such sequential instances are translated into the target sequence of the RNN model for generating image descriptions.
%
	\item [4)]
	{\bf SCST:Att2all}~\cite{5_}. This methods utilizes a ``soft'' top-down attention with context provided by a representation of a
partially-completed  sequence as context to weight visual-level features for captioning,
which is based on a reinforcement learning method with a self-critical sequence training strategy to train LSTM with expected sentence-level reward loss.
	\item [5)]
	{\bf Stack-Cap}~\cite{gu2018stack}. This method repeatedly produces the refined image description based on visual-level information with a \emph{coarse-to-fine} multi-stage framework, which incorporates the supervision information for addressing the vanishing gradient problem during training.
	\item [6)]
	{\bf Up-Down}~\cite{3_}. This method combines \emph{bottom-up} and \emph{top-down} attention mechanisms to calculate the region-level attentions for caption generation. To be specific, it is essentially based on a encoder-decoder architecture, and first makes use of bottom-up attention mechanism to obtain salient
image regions, and then the decoder employs two LSTM layers to compute the attentions at the level of objects and the salient regions with top-down attention mechanism for sentence generation.
\end{itemize}

\paratitle{Our method}.
Our proposed method is named \textsf{Stack-VS}, which is based on multi-stage architecture
with a sequence of decoder cells to repeatedly produce rich fine-gained image caption generation.
In particular, each decoder cell contains two LSTM-layers work interactively
to re-optimize attention weights on both visual-level feature
vectors and semantic- level attribute embeddings for generating a
fine-gained image caption.
%

\paratitle{Parameter Settings}.
To evaluate the quality of the image caption generation
results of different methods, we follow the evaluation strategy in~\cite{5_}.
That is,
we generate the vocabulary with a standard strategy to filter words,
namely, the words appearing less than $5$ times in frequency are filtered out.
After the processing, we obtained $8,791$ words in the final vocabulary.
For a fair empirical comparison, we use the same setting in ~\cite{3_} to extract visual-level feature (\ie $\matrix{V_0}$),
where $N_v = 36$ and each feature vector in $\matrix{V_0}$ is formed as a $2,048$-dimensional \emph{mean-pooled}  convolutional feature vector
to the corresponding salient region.
%
Similarly, we adopt multiple-instance learning(MIL)~\cite{2_,23_} model to obtain  the semantic attribute embeddings (\ie $\matrix{E_0}$),
where $N_e=20$ and each attribute embedding is formed as a $2,048$-dimensional vector converted by an embedding matrix that is randomly initialized and learned independently during training.

Our proposed model is implemented based on LSTM network,
and the parameters for our proposed method are empirically
set as follows:
(1) The number of hidden units in \emph{visual} attention model and the semantic model are set to $512$, respectively;
(2) We use adaptive moment estimation (Adam~\cite{kingma2014adam}) for optimization in the process of supervised cross-entropy training,
and the learning rate is set to $5e-4$, which is shrunk by $0.8$ for each $3$ epochs;
(3) The initial increasement rate of \emph{scheduled sampling} is set as $0.05$ for each $5$ epochs,
and its upper bound is set as $0.25$. After $30$ epochs, we make use of a \emph{reinforcement learning} based method to continue
to optimize it according to the learning rate, which is set by $5e-5$;
and (4) The batch size is set to $78$, and the maximum number of epochs is $100$, analogous to~\cite{gu2018stack}.

\setlength{\tabcolsep}{5pt}
\begin{table*}[!htbp]
	\centering
    \caption{Comparison results on MSCOCO on Karpathy's splits. The highest scores are highlighted in bold fold.}
    \label{tab:karpathy_splits}
	\renewcommand\arraystretch{1.3}
	\newcommand{\tabincell}[2]{
		\begin{tabular}{@{}#1@{}}#2\end{tabular}}
	\begin{tabular}{l|clclclclclclc|c|c|c|}
		\hline
		{\textbf{Method}}  &
		\tabincell{l}{\multicolumn{1}{c}{\textbf{B-1}}}  &
		\tabincell{l}{\multicolumn{1}{c}{\textbf{B-2}}}  &
		\tabincell{l}{\multicolumn{1}{c}{\textbf{B-3}}} &
		\tabincell{l}{\multicolumn{1}{c}{\textbf{B-4}}} &
		\tabincell{l}{\multicolumn{1}{c}{\textbf{METEOR}}} &
		\tabincell{l}{\multicolumn{1}{c}{\textbf{RougeL}}} &
		\tabincell{l}{\multicolumn{1}{c}{\textbf{CIDEr}}} &
		\tabincell{l}{\multicolumn{1}{c}{\textbf{SPICE}}} \\
		\hline
		Hard-Attention~\cite{1_}
		& 0.718  & \tabincell{l}{\multicolumn{1}{c}{0.504}}
		& 0.357  & \tabincell{l}{\multicolumn{1}{c}{0.250}}
		& 0.230 & \tabincell{l}{\multicolumn{1}{c}{-}}
		& -        &\tabincell{l}{\multicolumn{1}{c}{ -}} & \\
		
		Semantic-Attention~\cite{2_}
		& 0.709  & \tabincell{l}{\multicolumn{1}{c}{0.537}}
		& 0.402 & \tabincell{l}{\multicolumn{1}{c}{0.304}}
		& 0.243& \tabincell{l}{\multicolumn{1}{c}{-}}
		& -& \tabincell{l}{\multicolumn{1}{c}{-}} & \\

		MAT~\cite{6_}
		& 0.731  & \tabincell{l}{\multicolumn{1}{c}{0.567}}
		& 0.429 & \tabincell{l}{\multicolumn{1}{c}{0.323}}
		& 0.258 & \tabincell{l}{\multicolumn{1}{c}{0.541}}
		& 1.058 & \tabincell{l}{\multicolumn{1}{c}{0.189}} & \\ 		
		
		SCST:Att2all~\cite{5_}
		&  -        & \tabincell{l}{\multicolumn{1}{c}{-}}
		&  -        & \tabincell{l}{\multicolumn{1}{c}{0.342}}
		& 0.267 & \tabincell{l}{\multicolumn{1}{c}{0.557}}
		& 1.140 & \tabincell{l}{\multicolumn{1}{c}{-}} & \\
		
		Stack-Cap~\cite{gu2018stack}
		& 0.786  & \tabincell{l}{\multicolumn{1}{c}{0.625}}
		& 0.479  & \tabincell{l}{\multicolumn{1}{c}{0.361}}
		& 0.274 & \tabincell{l}{\multicolumn{1}{c}{0.569}}
		& 1.204 & \tabincell{l}{\multicolumn{1}{c}{0.209}} & \\
		
		$\mbox{Up-Down~[Cross-Entropy]}^{1}$~\cite{3_}
		& {\bf0.772} & \tabincell{l}{\multicolumn{1}{c}{-}}
		&  -        &  \tabincell{l}{\multicolumn{1}{c}{0.362}}
		& 0.270  & \tabincell{l}{\multicolumn{1}{c}{0.564}}
		& 1.135  & \tabincell{l}{\multicolumn{1}{c}{0.203}} & \\
		
		$\mbox{Up-Down~[CIDEr-Optimize]}^{2}$~\cite{3_}
		& {\bf0.798} & \tabincell{l}{\multicolumn{1}{c}{-}}
		&  -        &  \tabincell{l}{\multicolumn{1}{c}{0.363}}
		& 0.277  & \tabincell{l}{\multicolumn{1}{c}{0.569}}
		& 1.201  & \tabincell{l}{\multicolumn{1}{c}{0.214}} & \\
		\hline

		\textsf{Stack-VS} Attention [Cross-Entropy]~(Ours)
		& \tabincell{l}{\multicolumn{1}{c}{0.766}}
		& \tabincell{l}{\multicolumn{1}{c}{0.608}}
		& \tabincell{l}{\multicolumn{1}{c}{0.472}}
		& \tabincell{l}{\multicolumn{1}{c}{\bf0.366}}
		& \tabincell{l}{\multicolumn{1}{c}{\bf0.278}}
		& \tabincell{l}{\multicolumn{1}{c}{0.568}}
		& \tabincell{l}{\multicolumn{1}{c}{1.138}}
		& \tabincell{l}{\multicolumn{1}{c}{0.208}}   \\
		
		\textsf{Stack-VS} Attention [CIDEr-Optimize]~(Ours)
		& \tabincell{l}{\multicolumn{1}{c}{0.794}}
		& \tabincell{l}{\multicolumn{1}{c}{\bf0.636}}
		& \tabincell{l}{\multicolumn{1}{c}{\bf0.490}}
		& \tabincell{l}{\multicolumn{1}{c}{\bf0.372}}
		& \tabincell{l}{\multicolumn{1}{c}{\bf0.279}}
		& \tabincell{l}{\multicolumn{1}{c}{\bf0.577}}
		& \tabincell{l}{\multicolumn{1}{c}{\bf1.226}}
		& \tabincell{l}{\multicolumn{1}{c}{\bf0.216}}   \\
		\hline
	\end{tabular}
    \begin{tablenotes}
        \scriptsize
        \item[] {Note: ``-'' denotes the corresponding online results have not been reported.}
        \item[] {Note: $^{1}$~[Cross-Entropy]: indicates the model is trained with standard cross-entropy loss.}
        \item[] {Note: $^{2}$~[CIDEr-Optimize]: refers to the model is trained based on reinforcement learning and optimized for CIDEr score.}
    \end{tablenotes}
\end{table*}

\setlength{\tabcolsep}{10pt}
\begin{table*}[!t]
 \centering
  \caption{Comparison results on \textsf{MSCOCO} Caption Challenge 2015, using \textsf{MSCOCO} online evaluation server. The highest scores are highlighted in bold fold. All of results are evaluated based on c5 references.}
  \label{tab:MS_COCO_Test_serve}
 \renewcommand\arraystretch{1.3}
 \newcommand{\tabincell}[2]{
    \begin{tabular}{@{}#1@{}}#2\end{tabular}}
 \begin{tabular}{l|clclclclclclclclc|c|}
  \hline
  \multicolumn{8}{c}{\textbf{\textsf{MSCOCO} Captioning Challenge, $\textbf{\mbox{40,775}}$ Images (C5)}}\cr\cline{2-8}
  \hline
  {\textbf{Method}}  &
  \tabincell{l}{{\textbf{B-1}}} &
  \tabincell{l}{{\textbf{B-2}}}&
  \tabincell{l}{{\textbf{B-3}}} &
  \tabincell{l}{{\textbf{B-4}}} &
  \tabincell{l}{{\textbf{METEOR}}} &
  \tabincell{l}{{\textbf{RougeL}}}&
  \tabincell{l}{{\textbf{CIDEr}}} &\\
  \hline
  GoogleNIC~\cite{24_} &0.713&0.542&0.407&0.309&0.254&0.530&0.943&\\
  Hard-Attention~\cite{1_} &0.705&0.528&0.383&0.277&0.241&0.516&0.865&\\
  Semantic-Attention~\cite{2_} &0.731&0.565&0.424&0.316&0.250&0.535&0.943&\\
  MAT~\cite{6_} &0.734&0.568&0.427&0.320&0.258&0.540&1.029&\\
  Adaptive~\cite{18_} &0.748&0.584&0.444&0.336&0.264&0.550&1.042&\\
  Stack-Cap~[CIDEr-Optimize]~\cite{gu2018stack}&0.778 &0.616&0.468&0.349&0.270&0.562&1.148&\\
  Up-Down~[CIDEr-Optimize]~\cite{3_} &\bf{0.802}&\bf{0.641}&\bf{0.491}&0.369&0.276&0.571&1.179&\\\hline
  \textsf{Stack-VS} Attention~[CIDEr-Optimize]~(Ours) &0.790&0.634&0.489&\bf{0.372}&\bf{0.278}&\bf{0.575}&\bf{1.189}\\
  \hline
 \end{tabular}
\end{table*}

\subsection{Quantitative Analysis }
In this section, we evaluate the effectiveness of generating the image captions by our approach, in comparison to the
baseline methods on the test portion of the ``Karpath'' test splits.
For fair comparison, all of compared methods are merely trained on \textsf{MSCOCO} dataset,
and thus the models using external information will not be considered as candidates for comparison.
Table \ref{tab:karpathy_splits} and Table \ref{tab:MS_COCO_Test_serve} shows
\emph{BLEU} (\ie B-1, B-2, B-3, B-4), \emph{METEOR}, \emph{RougeL}, \emph{CIDEr} and \emph{SPICE} of each method.

From Table \ref{tab:karpathy_splits} and Table \ref{tab:MS_COCO_Test_serve}, we observe that:
\emph{First}, \emph{Stack-Cap} \cite{gu2018stack} performs better than \emph{Hard-Attention}, \emph{Semantic-Attention},
\emph{MAT} and \emph{SCST:Att2all} on all metrics.
For example, \emph{Stack-Cap} outperforms \emph{SCST:Att2all} by $5.56\%$, $2.62\%$, $2.15\%$
and $5.61\%$ in terms of \textbf{BLEU-4}, \textbf{METEOR}, \textbf{RougeL} and \textbf{CIDEr}, respectively.
This is because \emph{Stack-Cap} employs a sequence of decoder cells for repeatedly refining the image descriptions
while \emph{SCST:Att2all} is based on a simple one-pass attention architecture for training, which hardly generate rich and more
human-like captions.
The results demonstrate that using multi-stage structure is more effective than using one-pass.
\emph{Second}, \emph{Up-Down [CIDEr-Optimize]} outperforms \emph{Stack-Cap}
by $1.53\%$ and $2.39\%$ in terms of \textbf{BLEU-1} and \textbf{SPICE} respectively，
as \emph{Up-Down} computes the region-level attentions with a combination of \emph{top-down}
top-down attention mechanisms for captioning, which demonstrates the effectiveness of the combination architecture.
\emph{Third}, our proposed model \textsf{Stack-VS} consistently outperforms all baseline methods, and the improvements are statistically significant
on all metrics. For example, \textsf{Stack-VS} outperforms \emph{Stack-Cap}
by $3.05\%$, $1.82\%$, $1.41\%$, $1.83\%$ and $3.35\%$ in terms of \textbf{BLEU-4}, \textbf{METEOR}, \textbf{RougeL}, \textbf{CIDEr} and \textbf{SPICE}.
The reason might be due to two facts:
(1) The predicted words based on \emph{bottom-up} are a degree of depending on the \emph{visual-semantic} information of the input image,
and thus \textsf{Stack-VS} simultaneously handle visual-level and semantic-level information when generating fine-gained details like pre-positions;
(2) The generated captions essentially need to dependent on more \emph{visual-level} information when generating several semantic words such as \emph{noun} or \emph{adjective}.
However, \emph{Stack-Cap} solely relies on \emph{visual}-level information, which results in more errors while generating some semantic auxiliary words like preposition.
On the other side, our proposed model also outperforms \emph{Up-Down [CIDEr-Optimize]} by
$2.48\%$, $0.72\%$, $1.41\%$, $2.08\%$ and $0.93\%$ in terms of \textbf{BLEU-4}, \textbf{METEOR}, \textbf{RougeL}, \textbf{CIDEr} and \textbf{SPICE},
even though it also has a similar LSTM attention mechanism like ours.
The explanation is that \emph{Up-Down} solely relies on \emph{visual}-level information while neglecting the semantic-level one,
and meanwhile the a single stage based traing process also limits its capability in generating a fine gained caption.
In contrast, our proposed model is based on a \emph{coarse-to-fine} structure
with a sequence of decoder cells for repeatedly generating fine-gained caption in a stage-by-stage manner,
and at each time step, each decoder cell can simultaneously handle
both visual-level and semantic-level information for better capturing the interactions of
salient regions, and then assign different attention weights
for visual-level and semantic-level hidden states
for improving the accuracy and richness of the generated descriptions.


%
%
%



\begin{figure*}[!htbp]
	\centering	
    \includegraphics[width=2\columnwidth, angle=0]{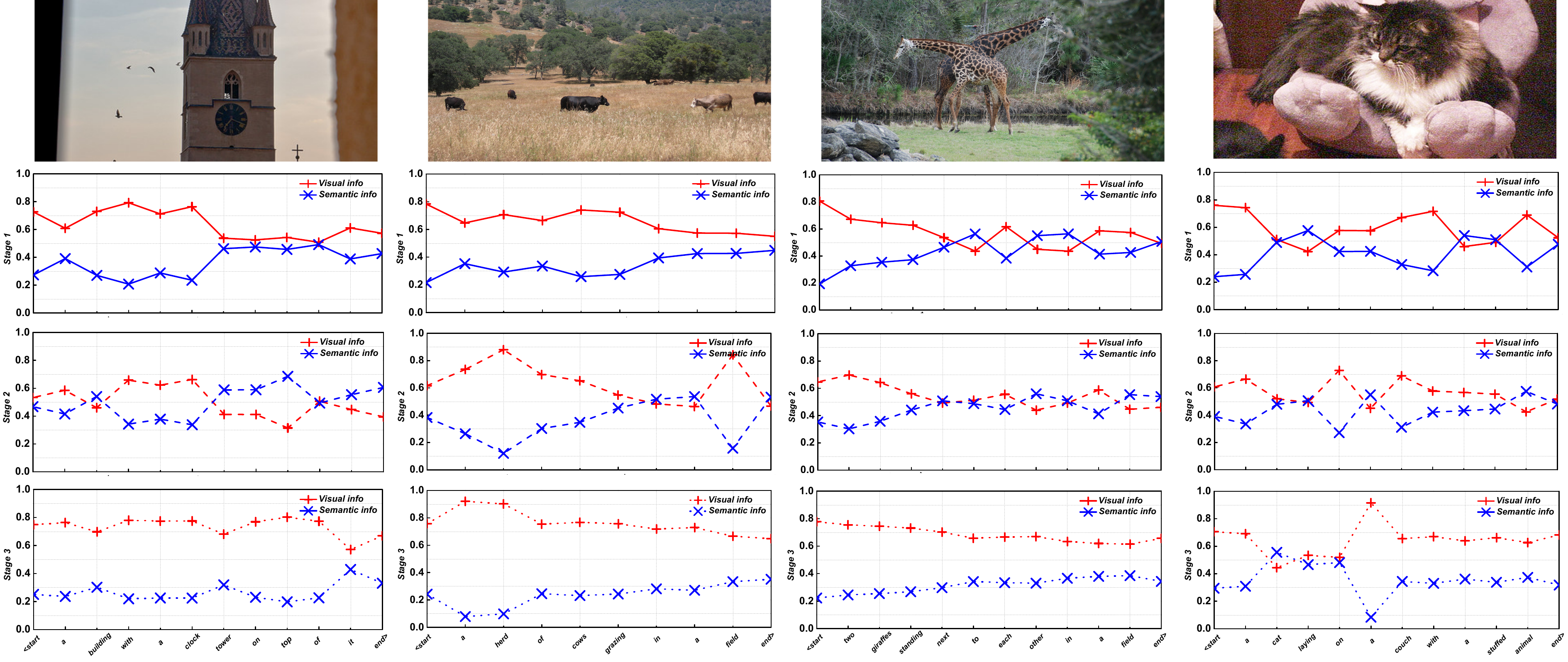}
	\caption{Examples of the proportion of the contribution of different level information (\ie \emph{visual}-level and \emph{semantic}-level) along with the image caption generation process at different stages.}
	\label{fig:Visual_vs_Semantic}
\end{figure*}
%
\begin{figure*}[!htbp]
	\centering	
	\includegraphics[width=2\columnwidth, angle=0]{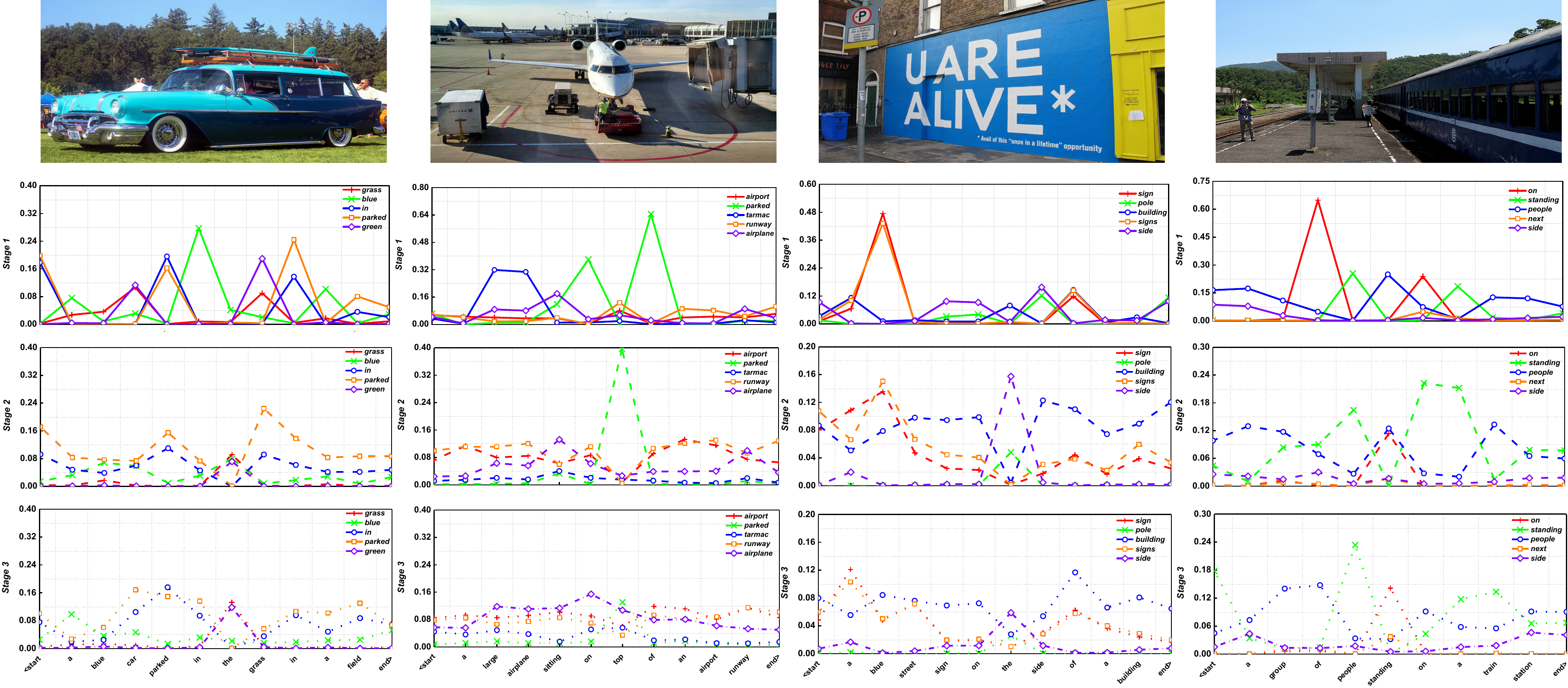}
	\caption{Examples of the process of learning semantic-level attention weight along with the caption generation at the different stages. X-axis presents the generated words, and Y-axis indicates the corresponding weight of each word.}
	\label{fig:Semantic_analysis}
\end{figure*}

\begin{figure*}[!t]
	\centering	
	\includegraphics[width=2\columnwidth, angle=0]{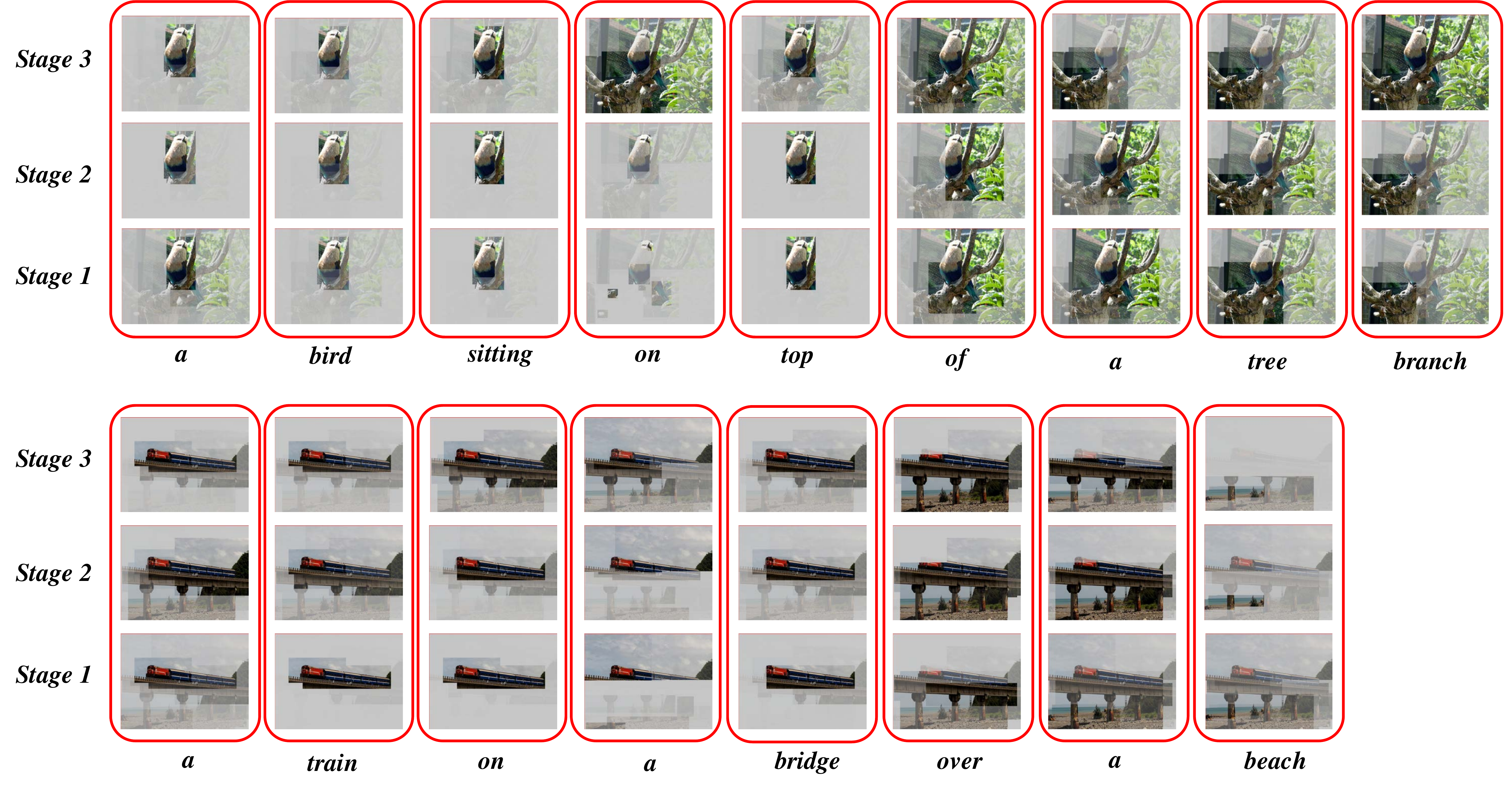}
	\caption{Examples of the process of learning visual-level attention weight along with the caption generation at the different stages. Specifically, the given sentences are generated by our proposed model and the change process of visual-level attention is shown in the corresponding images.}
	\label{fig:Visual_analysis}
\end{figure*}

\begin{figure*}[!t]
	\centering	
    \includegraphics[width=1.8\columnwidth, angle=0]{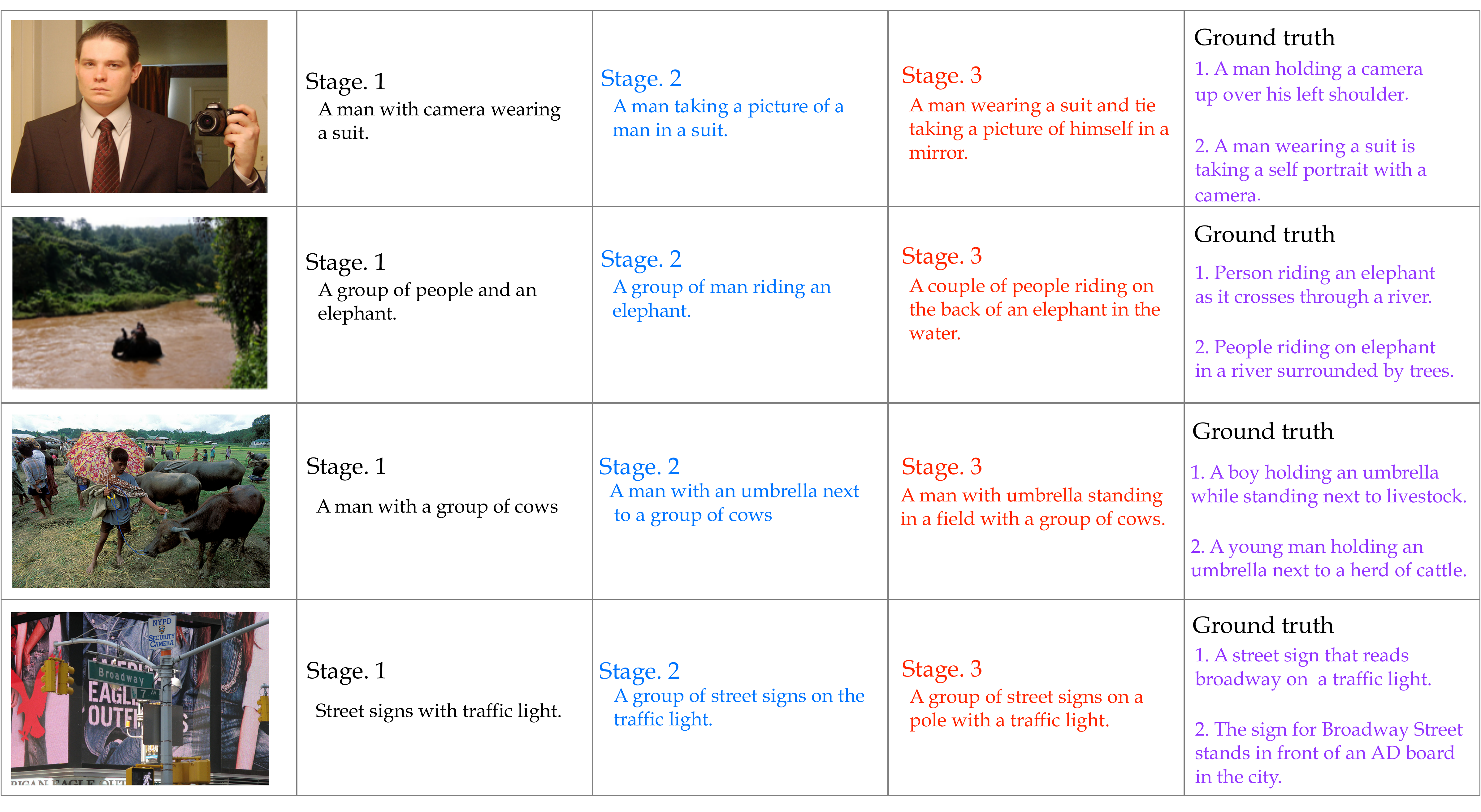}
    \caption{The graph at the top left shows the ratio of visual and semantic contribution. The graph at the top right show the semantic attention distribution on the five words with the largest average weight. The graph at the bottom show the visual attention distribution on salient regions.}
	\label{fig:stage_results}
\end{figure*}

\subsection{Qualitative Analysis}
In this section, we conduct an in-depth analysis on our proposed model to demonstrate the superiority of generating better description within the proposed multi-stage
\emph{visual-semantic} attention based framework.

\paratitle{Different Impacts of Visual-level and Semantic-level Information}.
From Fig.~\ref{fig:Visual_vs_Semantic}, we can observe that our proposed model is capable of adaptively leveraging the different contributions derived from \emph{visual}-level and \emph{semantic}-level information for captioning.
More concretely, the proportion of the attended weights of \emph{semantic}-level information increases when generating prepositions (\eg ``in'', ``front''
or ``of''), however generating some nouns-related objects (\eg ``people'', ``rides'',``trains'',``bicycles'')
would make the curve drops to some extent, which is able to adaptively make use of both \emph{visual}-level and \emph{semantic}-level
information for image caption generation.

\paratitle{Impact of Semantic-level Attention}. To qualitatively evaluate the impact of \emph{semantic}-level attention in our proposed model,  top-$5$ words with the highest attention scores of each input image are selected for analysis. The observations from Fig.~\ref{fig:Semantic_analysis} are as follows\footnote{Note that we use different types of lines to denote different stages.}, the branch of \emph{semantic}-level attention in \textsf{Stack-VS}: (1) Has a good capability of gradually refining the attention scores to output the most appropriate next prediction word via the  multi-stage structure. Without loss of generality, we take the first image as example.
After processed, the word ``grass'' and ``parked'' in the final stage (\eg stage-3) obtain the highest attention scores at the previous time step, and thus the  prediction words at the current time step are correctly outputted;
and (2) Filters out noises and adaptively learn the linguistic pattern to assign the highly-relevant word pair with the highest scores, such as ``green'' and ``grass'', the attention scores are far from each other in stage-1, however they are assigned the top-2 scores in stage-3 after refining, as the adjective ``green''
is the most appropriate adjective to describe the noun ``grass''.
%

\paratitle{Impact of Visual-level Attention}. To demonstrate using \emph{visual}-level attention can generate more plentiful image captions,
we visualize the assignment process of visual-level attention weight at each time step for an in-depth analysis.
As shown in Fig.~\ref{fig:Visual_analysis}, we observe that our proposed \textsf{Stack-VS} model can gradually filter out noises and the distribution of the visual-level attentions at previous time step are concentrated to the corresponding regions that are highly relevant to the next prediction words, and the learned assignments are correctly consistent with human intuition.
We take the first image as example, the attention visualizations of the objects like ``bird'', ``train'' or ``beach'', and the distribution of \emph{visual}-level attention is highly pinpointed to the corresponding salient regions, which indicates that learning the \emph{visual}-level attention is effective for generating the high quality image captions.


\paratitle{Impact of Stacked Refinement}.
Here, we mainly focus on the analysis of the impact of stacked refinement in our proposed visual-semantic attention based multi-stage architecture.
In fact, the caption generated in the coarse stage is proper for understanding the input image, however the accuracy and richness of the generated
description is insufficient.
We can take the images listed in Fig.~\ref{fig:stage_results} as example for an in-depth analysis.
From the figure, we can observed that the ultimately predicted words are obtained by superimposing the \emph{visual}-level and \emph{semantic}-level
information to refine the description at each stage for generating a more fine-gained caption.
Indeed, the model at the next stage re-assigns the \emph{visual-semantic} attention weights for fined-tuning weights to repeatedly
output several auxiliary words to infer the semantic details of the given image using the information from the previous stages.
Hence, actually the sum of semantic-level weights at stage-3 is lower than the previous ones,
as the attention weights are not strictly assigned to the limited scope of candidate words.
Without loss of generality, we take the first image as example.
The first stage is good enough to detect the correct objects bridged with several simple prepositions for captioning.
At the next stage, the process of the stacked refinement is to capture the relations among objects to form the skeleton
of the caption, and then refine several words to supplement the details to generate more natural and human-like captions
through multiple stages.
For example, in the final caption, which utilizes `` wearing a suit '' to replace of ``in a suit'', and
add some details to describe ``a man'', such as ``wearing ... and tie'', ``in a mirror''.


%

\section{Conclusion}
\label{sec:con}
In this paper, we have proposed a \emph{visual-semantic} attention based multi-stage framework for image caption generation problem. Specifically, the proposed model is based on a combination of \emph{top-down} and \emph{bottom-up} architecture with a sequence of decoder cells,  in which each decode cell is re-optimized for the hidden states of each decoder cell linked one-by-one at each time step. In particular, differ from previous studies that merely feed the \emph{visual}-level or \emph{semantic}-level features alone into the decoder to generate descriptions, our proposed model simultaneously feed both of such information into each decoder cell
to work interactively to repeatedly refining the  weights via two \textsf{LSTM} based layers. Experiments conducted on the popular dataset \textsf{MSCOCO} demonstrate that our proposed model demonstrate the comparable performance over the state-of-the-art methods using ensemble on the online MSCOCO test server. For future work, we plan to investigate the following issues: (1) Incorporating natural language inference to generate more reasonable captions; and (2) Trying different architectures for captioning, such as graph convolution network.


	
\begin{IEEEbiography}{Wei Wei}
	received the PhD degree from Huazhong University of Science and Technology, Wuhan, China, in 2012. He is currently an associate professor with the School Computer Science and Technology, Huazhong University of Science and Technology. He was a research fellow with Nanyang Technological University, Singapore, and Singapore Management University, Singapore. His current research interests include computer vision, natural language processing, information retrieval, data mining, and social computing.
\end{IEEEbiography}
\begin{IEEEbiography}{Ling Cheng}
	received the master degree from Fudan University, Shanghai, China, in 2019. He is currently a PhD student at Singapore Management University, Singapore. His research interests include computer vision and deep learning.
\end{IEEEbiography}
%

\begin{IEEEbiography}{Xian-Ling Mao}
received the PhD degree from Peking University, in 2013. He is currently an associate professor of computer science with the Beijing Institute of Technology. He works in the fields of machine learning and information retrieval. His current research interests include topic modeling, learning to hashing, and question answering.

Dr. Mao is a member of the IEEE Computer Society and a member of the Association for Computing Machinery (ACM).
\end{IEEEbiography}

\begin{IEEEbiography}{Guangyou Zhou}
received the PhD degree from National Laboratory of Pattern Recognition (NLPR), Institute of Automation, Chinese Academy of Sciences (IACAS) in 2013. Currently, he worked as a Professor at the School of Computer, Central China Normal University. His research interests include natural language processing and information retrieval. He has published more than 30 papers in the related fields.
\end{IEEEbiography}

\begin{IEEEbiography}{Feida Zhu}
is an associate professor at Singapore Management University (SMU), Singapore.
He received his Ph.D. degree from the University of Illinois at Urbana-Champaign (UIUC) in 2009.
His research interests include large-scale data mining, text mining, graph/network mining and social network analysis.
\end{IEEEbiography}

\end{document}